\pdfoutput=1

\documentclass[11pt]{article}

\usepackage{emnlp2021}

\usepackage{times}
\usepackage{latexsym}

\usepackage[T1]{fontenc}

\usepackage[utf8]{inputenc}

\usepackage{microtype}

\usepackage{color}
\usepackage{amsmath}

\usepackage{graphicx}
\usepackage{natbib}
\usepackage{caption}
\usepackage{amsfonts}
\usepackage{amssymb}
\usepackage{multirow}
\usepackage{adjustbox}
\usepackage[english]{babel} 
\usepackage[ruled,linesnumbered]{algorithm2e}
\usepackage{bbding}
\usepackage{CJKutf8}
\usepackage{booktabs}
\usepackage{listings}

\usepackage{makecell}

\SetKwInOut{Parameter}{Parameters}

\title{DataCLUE: A Benchmark Suite for Data-centric NLP}

\author{
\begin{tabular}{c} 
Liang Xu,~~ Jiacheng Liu\thanks{$^\ast$ Corresponding author. E-mail: liujiacheng@sjtu.edu.cn},~~ Xiang Pan,~~ Xiaojing Lu,~~ Xiaofeng Hou \\
 \\
  CLUE team \\
   {\tt \ CLUE@CLUEbenchmarks.com} 
   \end{tabular}
}

\begin{document}
\maketitle
\begin{abstract}







%
%

Data-centric AI has recently proven to be more effective and high-performance, while traditional model-centric AI delivers fewer and fewer benefits.
It emphasizes improving the quality of datasets to achieve better model performance. 
This field has significant potential because of its great practicability and getting more and more attention. 
However, we have not seen significant research progress in this field, especially in NLP. Therefore, we propose DataCLUE, which is the first Data-Centric benchmark applied in NLP field. We also provide three simple but effective baselines to foster research in this field (improve Macro-F1 up to 5.7\% point). In addition, we conduct comprehensive experiments with human annotators and show the hardness of DataCLUE.
Also, we try an advanced method: the forgetting informed bootstrapping label correction method. All the resources related to DataCLUE, including dataset, toolkit, leaderboard, and baselines, is available online.\footnote{\url{https://github.com/CLUEbenchmark/DataCLUE}}




\end{abstract}

\section{Introduction}\label{Introduction}




Existing works in AI are mostly model-centric and try to improve the performance through creating a new model. 
After years of development, we have many models that perform well in different tasks,
such as BERT \cite{bert-2019} in natural language processing and ResNet \cite{he2016deep} in computer vision. 
Furthermore, these models can be easily obtained from open-source websites such as GitHub. 
Although new models are proposed every day, many of these may not boost performance in a large margin or rarely be used~\cite{dacrema2019we, lv2021we}. 
At the same time, a lot of practical experience in the industry shows that focusing on optimizing data instead of models can often achieve better results~\cite{MLOps}.

 \begin{figure}[t]
	\includegraphics[width=0.5\textwidth]{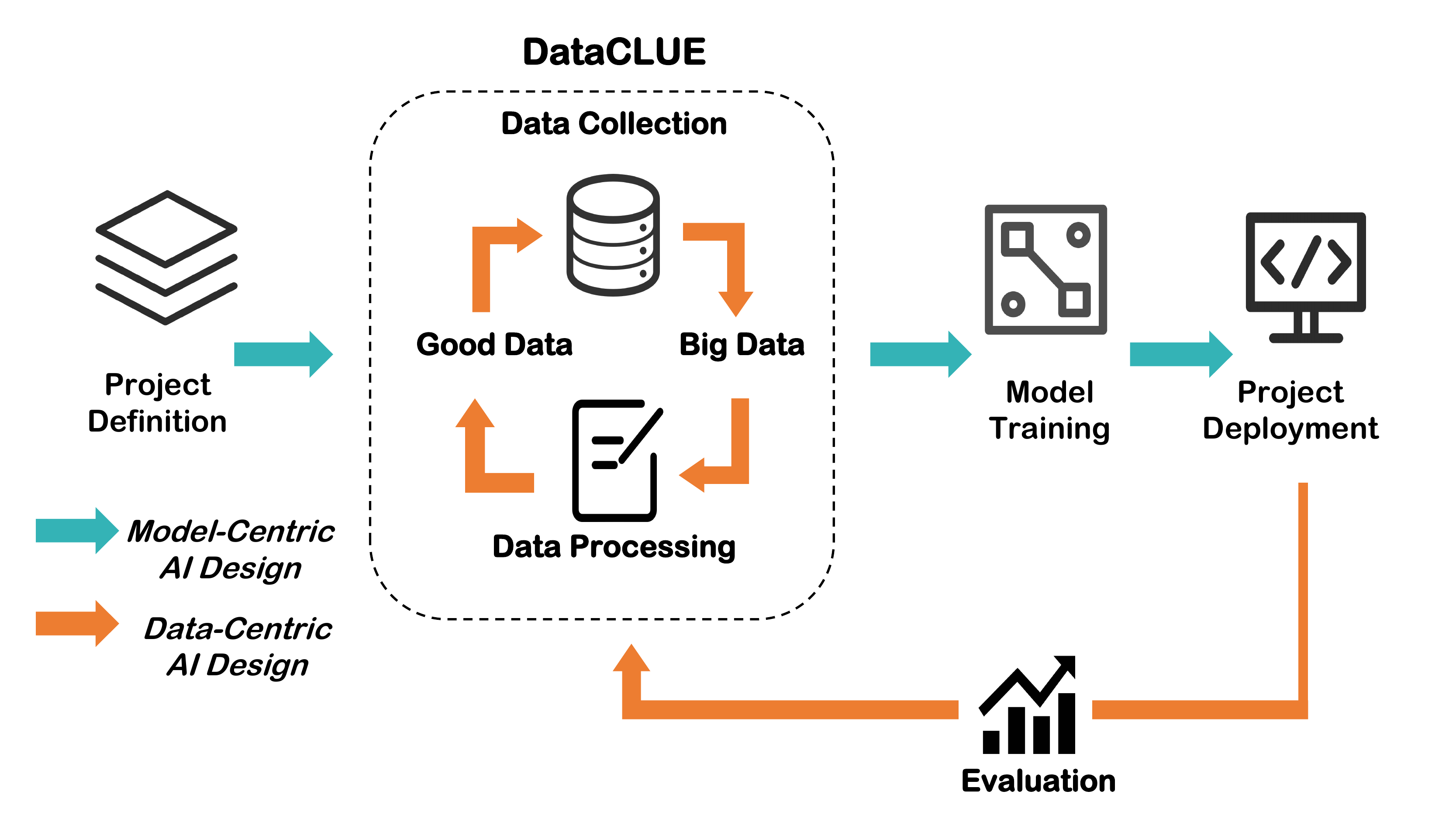}
	\caption{
		Overview of DataCLUE
	}
	\label{fig:overview}
\end{figure}


The emerging field of Data-Centric Artificial Intelligence (AI) is expected to deliver a set of techniques for dataset optimization, thereby enabling models like deep neural networks to be effectively trained using smaller datasets.
Although under the umbrella of data-centric AI, there are already many related researches that are developing independently, e.g., data augmentation~\cite{wei-zou-2019-eda}, data cleaning~\cite{Tong2014CrowdCleanerDC} etc. However, this field still lacks some general benchmark tasks, which also severely restricts the further development of data-centric AI.

In this paper, we propose DataCLUE, a data-centric AI benchmark suite for data-centric NLP (Figure \ref{fig:overview}). 
DataCLUE is based on the CLUE benchmark~\cite{clue} and combined with the typical characteristics of data-centric AI. 
It provide a general evaluation platform for data-centric AI. 
In DataCLUE, we select several representative tasks together with the evaluation metrics.
We also provide an online leaderboard for researchers to submit and compare their results with others.
To help more researchers get start with data-centric AI, we also provide several simple and effective baselines with open source code. 
Last but not least, we conduct comprehensive experiments to verify the effectiveness of these simple baselines.
Meanwhile, we try to use human annotation and some other STOA data curation methods in DataCLUE.
The results show that tasks in DataCLUE are very challenging, and cannot be solved well by simply using manual annotation or existing methods.

In summary, our contribution is three-fold:
\begin{itemize}
 \item We introduce DataCLUE, the first data-centric NLP benchmark suite. 
 It provides a wide range of tasks from different domains.
 
 \item We propose three simple baselines, and conduct comprehensive experiments on them. The results show they are very effective that can beat expensive human annotators and one of the state-of-the-art data optimization method.

 \item We open-source the tasks, baselines, toolkit, and developed a public leaderboard, to help advance research in data-centric AI. 
 
 
 \end{itemize}

\section{Related Work}\label{RelatedWork}

\subsection{Data-Centric AI}
Data-centric AI proposes a new perspective on improving the performance of AI systems. It focuses on making the dataset better instead of spending much time in models~\cite{MLOps}.
It has gained much attention recently, \cite{hajij_2021_DataCentricAIRequires} claims that data notion behind Data-Centric AI and we need a clear interest definition of the final task.
Under the umbrella of methods in data-centric AI, there are many ways to improve the quality of datasets. 
In the collection phase, it's a common practice to gather duplicate labels from different workers and aggregate them to get the accurate estimation~\cite{Zheng2017TruthII}.
Upon obtaining the dataset, a human-in-the-loop approach can be used to clean the data~\cite{Tong2014CrowdCleanerDC}.
During the learning procedure, there are also many approaches that can improve the data quality by augmenting the dataset with different transformations~\cite{wei-zou-2019-eda}. 
There are also methods that train a model with noisy data and utilize the prediction results to find labeling errors~\cite{Li2019LearningTL}. 


\subsection{Related Benchmarks}
As an emerging field, the are few works in benchmarking Data-centric AI, especially in the NLP domain.
The most related one is the data-centric AI competition held by Andrew Ng~\cite{ng_data_centric_ai_competition} which is an image classification dataset.
Besides, there are some model-centric benchmarks such as the CLUE~\cite{clue}, GLUE~\cite{wang2018glue}.
However, these benchmarks cannot be used to evaluate the performance of data-centric approaches.


\section{DataCLUE Overview}\label{Tasks}
\begin{table*}[t]
\footnotesize
	\begin{adjustbox}{width=1\textwidth}
\begin{tabular}{llllllllll}\toprule
Corpus & Train & Dev & Test Pub. & Test Pri. & Classes & Unlabeled & Task & Metric & Domain \\ \midrule
CIC & 10000 & 2000 & 2000 & $\ge$3000 & 118 & 14790 & \begin{tabular}[c]{@{}l@{}}Intent\\ Classification\end{tabular} & Macro-F1  & E-Commrce \\
IFLYTEK & 12133 & 2599 & 2000 & $\ge$3000 & 119 & 10000 & \begin{tabular}[c]{@{}l@{}}Short Text\\ Classification\end{tabular} & Macro-F1  &  App Descriptions \\
TNEWS & 53360 & 10000 & 2000 & $\ge$3000 & 15 & 10000 & \begin{tabular}[c]{@{}l@{}} Short Text\\ Classification\end{tabular} & Macro-F1  &  News Title \\
\bottomrule
 \hline
\end{tabular}
	\end{adjustbox}
\caption{Tasks description and statistics. Our experiments report on CIC. We are working on IFLYTEK and TNEWS. }
\label{tab:tasks:description}
\end{table*}



\subsection{Task Definition}
Our tasks aim at evaluating methods that can improve the data set to boost the model's performance. 
In these tasks, participants need to improve the data set under the task to improve the final performance of the model. 
They can modify the training set and validation set, re-split the training set and validation set, or add data by non-crawler methods. 
The modification can be done by algorithms or programs or in combination with manual methods. 
Finally, participants need to use the modified training set and validation set to train a fixed model, and evaluate the model on the public test set (or submit it to the leaderboard to get a ranking).

\subsection{Task Building Process} 
Data-centric AI is more suitable for the industrial applications because it considers data quality issues that are common in real-world.

To this end, we collected and constructed several data sets that fit this situation. 
For data that has no noise, we use the following process to construct noise that conforms to the real-world.
Firstly, we train a classification model with additional data to predict the probability distribution of labels for each data in the training set. Then, we randomly select 40\% of the data and randomly use one of the following strategies to assign a new label.
 \begin{itemize}
     \item \textit{Random Flip.} We uniform randomly replace the label with one of the label in the label list. 
     
     \item \textit{Hard Flip.} 
     To construct hard noise data for DataCLUE, 
     we use a model to predict the data, then for each data we define the top five most likely labels predicted by the model as \textbf{similar labels}. Then we randomly replace the label with the similar label.
 \end{itemize}
 
 After the above steps, we remove some easy data which can be identified by simple keywords or regularities to increase the task difficulty.

\subsection{Characteristics of Task}
In summary, the tasks in DataCLUE have the following characteristic, 
\begin{itemize}
 \item \textit{Representative.} In order to truly reflect the characteristics of data in real applications. We choose the
 training and validation set so that they have a high proportion of noisy. 
 E.g., for CIC more than 1/3 and 1/5 of the data in the training set and validation set may have the problem of labeling errors. 

 \item \textit{Challenging.}  The data are diverse and unevenly distributed. There are more than 100 classes, and some classes are easy to be confused between each other. 
 Meanwhile, the dataset is seriously unbalanced, making it more difficult to learn a good model.
 
 \item 
 \textit{Resource Friendly.}
 We set the model to be light weighted, so that the experimental cycle can be fast with common computing infrastructures. A single experiment in a GPU environment can be completed in about 4 minutes. 

 \item \textit{Academic Friendly.} We release a public test set so that participants can evaluate their results independently and replicate our experiments.
\end{itemize}

\subsection{Dataset Description}
At this time, we mainly focus on the classification task. We collected and processed three datasetd, the description and statistics of them are listed in Table~\ref{tab:tasks:description}.  We also plan to release more datasets in the near future.

We give some examples from CIC in Table~\ref{tab:examples}.

\begin{CJK}{UTF8}{gbsn}
\begin{table*}[htbp]
\footnotesize
\centering
\scalebox{1}{
\begin{tabular}{p{0.005\textwidth}p{0.93\textwidth}}
 \toprule
 
 \parbox[t]{1mm}{\multirow{2}{*}{\rotatebox[origin=c]{90}{{\textbf{CIC}}}}} &
\textbf{sentence:} 
你们一般情况发什么快递呢？
\\ &
\textbf{sentence (en):}
\textit{
What kind of express do you usually send?
}
\\ & \textbf{label:} \texttt{55(买家咨询发什么快递)} 
\\ & \textbf{label(en):} \texttt{55(Buyers consult what courier to send)} 
~\\
\midrule

  \parbox[t]{1mm}{\multirow{2}{*}{\rotatebox[origin=c]{90}{{\textbf{CIC}}}}} &
\textbf{sentence:} 
我想改地址和电话
\\ &
\textbf{sentence (en):}
\textit{
I want to change my address and phone number. 
}
\\ & \textbf{label:} \texttt{49(买家要求修改收件信息)} 
\\ & \textbf{label(en):} \texttt{49(The buyer requests to modify the receiving information)} 
~\\
\midrule

  \parbox[t]{1mm}{\multirow{2}{*}{\rotatebox[origin=c]{90}{{\textbf{CIC}}}}} &
\textbf{sentence:} 
有现货的明天能发货吗
\\ &
\textbf{sentence (en):}
\textit{
Can you deliver tomorrow if there is stock?
}
\\ & \textbf{label:} \texttt{54(买家咨询发货时间)} 
\\ & \textbf{label(en):} \texttt{54(The buyer inquires about the delivery time)} 
~\\
\midrule

\end{tabular}
}
\caption{Development set examples from the task in DataCLUE. \textbf{Bold} text represents part of the example format for each task. Chinese text is part of the model input, and the corresponding text in \textit{italics} is the English version translated from that.}

\label{tab:examples}
\end{table*}
\end{CJK}

\begin{lstlisting}[language=Python, caption={\textit{dckit} usage example} ,captionpos=b]
from dckit.utils import read_datasets,
                    random_split_data
from dckit.evaluate import evaluate

data = read_datasets()
# TODO do your transformation here.
data = example_transform(data)
random_split_data(data)
f1 = evaluate()
\end{lstlisting}

\subsection{DataCLUE Toolkit}
In order to further simplify the process of researchers using DataCLUE, we developed the DataCLUE Toolkit (\textit{dckit}). It contains the data reading, result generation, result validation and other functions. By using \textit{dckit}, research can implement a new data-centric methods with very few lines of code. As an example, you can see Listing 1.

\section{Baselines}\label{baseline}

Beside the DataCLUE benchmark, we propose several baseline methods from different perspectives. These methods aims to give a starting point for the newcomers, rather than achieve STOA. 
We broadly classify these methods into three groups based on the point they focus on, namely the label, data and label definition. 
All the baselines are also available in the DataCLUE repository.

\subsection{Mislabeled Deletion (y)}
The first strategy is target on the noisy labels.
Just like most real-world datasets, the datasets in DataCLUE have high proportion of noise. It is very intuitive to think of using an algorithm to filter out potentially incorrect labels for correction or discard. 

In this paper, we propose to use a pre-trained model that fine-tuned in DataCLUE to predict the label.
Since we need to get the estimation of each data points, we introduce a cross-validation-based approach. Consider we have the original data as $\mathbf{D}$, we firstly split it into $k$ part of each size, namely $\mathbf{D}_1, \mathbf{D}_2, \cdots, \mathbf{D}_k$. Then, for each iteration $i$, we train a model with all the splits except for $\mathbf{D}_i$, and then use the model to predict the label probability of $p$.
After cross-validation, the probability distribution predicted by the model is used to calculate entropy (representing the uncertainty of prediction). Namely, for data sample $x$ the error probability is calculate as,
\begin{equation}
    e_{x}=\sum_{i=0}^k p_{x}^i\log p_{x}^i
\end{equation}
where $p_x^i$ is the predict probability for sample $x$ belongs to class $i$. 

Finally, the dataset will then be sorted and remove the samples most likely to be mislabeled. 

  


\subsection{Data Augmentation (x)}
Then, we turn to optimizing the data (x).
Consider the datasets in DataCLUE may not be sufficient for the classification of many classes. We adopt the commonly used data augmentation techniques~\cite{wei-zou-2019-eda}. Specifically, we augment each data point by replacing synonym, random insert words and random delete words. 
This can effectively enhance the size of the dataset, which is very important for large model training.

\subsection{Definition Augmentation (metadata)}
 Finally, we turn our attention to label definition. Since most NLP datasets have a definition about what this class is. We can use this label definition to create more data. We can assign label definition as a sentence for the dataset if it lacks label definition.
 In this paper, we consider apply the data augmentation techniques to the label definitions to get more concrete data.
 Intuitively, if an input sentence is closer to the definition of the label, the model is easier to learn from it.


\section{Experiments}\label{experiments}
In this section we firstly describe the implementation details and then present the results of the different algorithms on the dataset. 

\subsection{Implementation Details}\label{sec:implementation}
\begin{figure}[t]
	\includegraphics[width=0.5\textwidth]{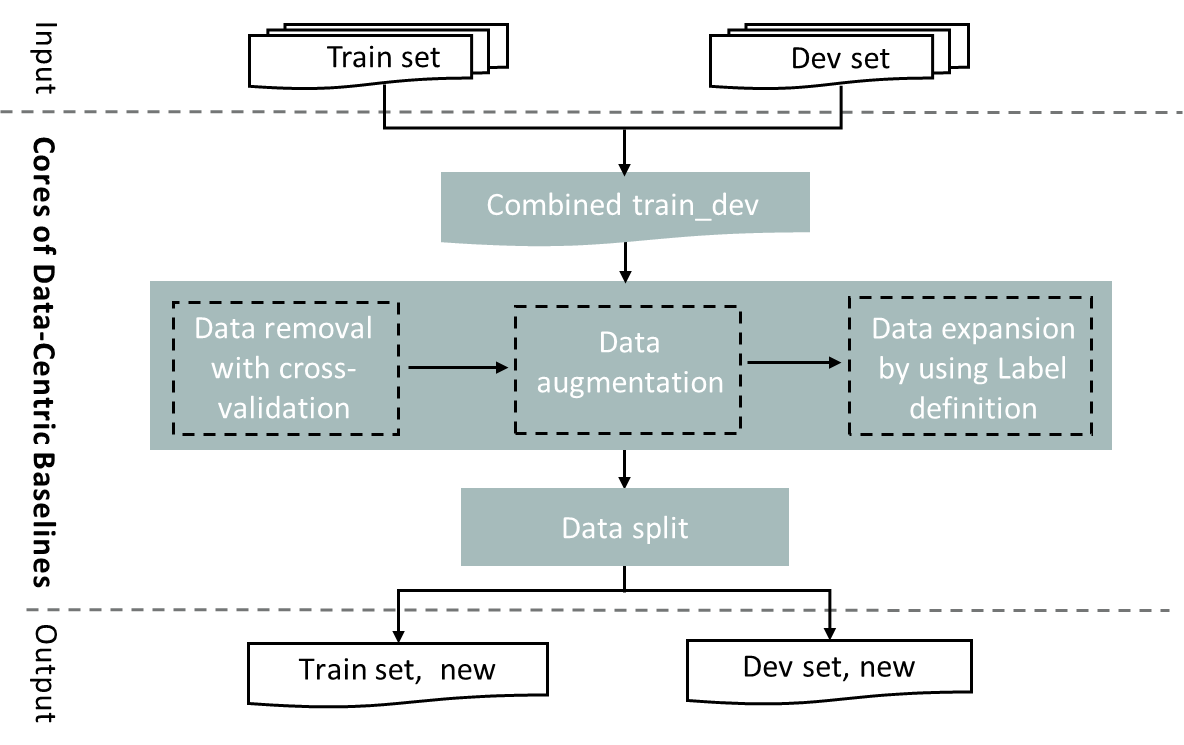}
	\caption{Training Procedure of the Baseline}
	\label{fig:training_precedure}
\end{figure}

We now describe implementation details for each component in DataCLUE.
For all the baselines, we firstly merge the training set and validation set. Then, we adopt the baseline methods as described in Section \ref{baseline}.
Finally, we split the processed dataset into training set and validation set using stratified random splits. Check Figure \ref{fig:training_precedure} for more detail.





In the evaluation process, we use RoBERTa-3L-wwm~\citep{cui-etal-2020-revisiting}\footnote{\urlstyle{same}\url{https://github.com/ymcui/Chinese-BERT-wwm}} as the default pre-training model. This is a three-layer model, which performance is close to the full-scale model and the training speed and efficiency are much better.
As for the parameters, we set sequence length as 32, epoch as 15, batch-size as 64, and learning-rate as 2e-5.
The results are calculated based on the public test set.
As this is a unbalanced classification task, we use Marco-F1 as the primary evaluation metric.

\subsection{Experiment Results}
In this subsection, we conduct experiments to analyze the performance of baseline methods in DataCLUE. We run all the experiments 5 times and report the average results. 

\begin{table}[htbp]
\footnotesize
\begin{tabular}{llllll}\toprule
Delete Num. & 0 & 100 & 500 & 1000 & 2000 \\ \midrule
Marco-F1 & 72.78 & \textbf{73.32} & 72.52 & 73.29 & 69.41\\
\bottomrule
 \hline
\end{tabular}
\caption{Analysis of Mislabeled Deletion.}
\label{tab:baseline:drop}
\end{table}
\subsubsection{Parameter analysis}
Firstly, We analyzed the impact of the number of discarded samples on performance. The results are shown in Table \ref{tab:baseline:drop}. As we can see delete the mis-labeled data can improve the performance. Except for discarding the first 2000 data points with highest entropy, all the other parameters have gain different improvement. 
We believe this is because by discarding the most likely incorrectly labeled data, we can effectively improve the signal-to-noise ratio of the data set, thereby obtaining better performance. However, discarding too much data may affect the final training effect. From our experimental results, discarding the first 100 data is the best choice.

\begin{table}[htbp]
\footnotesize
\begin{tabular}{llllll}\toprule
Aug. Num. & 0 & 1 & 3 & 5 & 10 \\ \midrule
Marco-F1 & 72.78 & 73.88 & \textbf{74.62} & 73.63 & 66.94\\
\bottomrule
 \hline
\end{tabular}
\caption{Analysis of Data Argumentation.}
\label{tab:baseline:aug}
\end{table}

Then, we analyzed the impact of different degrees of data enhancement on the final performance. The results are shown in Table \ref{tab:baseline:aug}. Similarly, we can find that data enhancement methods can effectively improve the effect of the model. At the same time, doing too much data enhancement may bring adverse effects (such as ten times enhancement or even no enhancement). This is because excessive enhancement may result in the generation of many samples that deviate from the original intention, thereby affecting the performance of the model. From our experimental results, choosing triple enhancement can get the best performance.

\begin{table}[htbp]
\footnotesize
\begin{tabular}{llllll}\toprule
Label Num. & None & 1 & 10 & 50 & 100 \\ \midrule
Marco-F1  & 72.78 & 73.82 & 76.38 & \textbf{78.22} & 77.97\\
\bottomrule
 \hline
\end{tabular}
\caption{Analysis of Definition Augmentation.}
\label{tab:baseline:text}
\end{table} 
Finally, we analyzed the impact of using label definitions on performance in Table \ref{tab:baseline:text}. Here, we use ``None'' to denote we do not use the label definitions.
We can see that using label definitions can greatly improve model performance. By directly introducing label definitions, we can achieve an Macro-F1 of 73.82, which has tasted the best performance of deleting mislabeled data. Through label augmentation we can get a higher improvement, and finally reach 78.22 Macro-F1. The use of augmentation here also has a similar phenomenon to the previous one, that is, too much data augmentation is meaningless. According to our experiment, a 50-fold augmentation is an optimal choice.

\begin{table}[htbp]
\centering
\footnotesize
\begin{tabular}{cccc}\toprule
Noisy Delete & Data Aug. & Label Aug. & Marco-F1 \\ \midrule
 & & &  72.78\\
\checkmark  & & &  73.32(+0.54)\\
  & \checkmark& &  74.62(+1.84)\\
  & & \checkmark&  78.22(+5.44)\\
\checkmark  & \checkmark& &  74.78(+2.00)\\
\checkmark  & & \checkmark&  \textbf{78.49}(+5.71)\\
  & \checkmark  & \checkmark&  75.82(+3.04)\\
\checkmark  &\checkmark &\checkmark &  76.23(+3.45)\\
\bottomrule
 \hline
\end{tabular}
\caption{Combination of different strategies.}
\label{tab:baseline:comb}
\end{table}

\subsubsection{Combination Analysis}
In this part, we try to figure out if we can use these baseline methods together to get better performance. The results are shown in Table \ref{tab:baseline:comb}. Through this comparison, we can see that the combination of the three strategies can effectively improve the performance of the final model. Especially, the combination of noisy label delete and label aggregation can achieve the best performance of 0.7849. Besides, we find that the label augmentation and the data augmentation methods looks incompatible. This may due to the over-augmentation creates too much noise.

In summary,  the three proposed baseline methods can achieve excellent performance in Data-centric AI applications. 
In the following parts, we will analyze the performance of other methods to show the effectiveness of this simple baselines.

\section{Label Correction Methods}
In this section, we discussed the effectiveness of human additional annotation and a label correction method based on the distribution of the forgetting event.

\subsection{Human-in-the-loop Pipeline}
Using human to annotate or verify data in the data-centric AI pipeline is thought to be an effective method. 
In this section, we try to figure out the effectiveness of this method in DataCLUE.

\subsubsection{Annotation Settings}\label{sec:anno_process}
We use a commercial annotation service to conduct experiments in this section. 
The overall annotation procedure includes three steps,
\begin{itemize}
    \item Firstly, the expert annotators will conduct trial labeling, combined with a small number of samples and label definition files to understand the difficulty and completion of labeling tasks and labeling. 
    We also set the requirement that human annotation accuracy must above 0.92.
    
    \item Secondly, the company will estimate the annotation price and the completion time according to the performance on the trail labeling task. 
    In our experiments the cost of labeling a single data sample is \yen 0.35 (or \$0.06). And the completion time of the labeling task (consists 2000 data points) is 10 hours.
    
    \item Finally, the expert annotators will instruct the annotators. 
    The annotators will also need time to become familiar with the definition of labels and learn during the annotation process. 
    After labeling, expert annotators will re-inspection the task using spot-check. When the results meet the requirements of desired accuracy, the labeled data will be send to us.
\end{itemize}

Using this standardized labeling process can ensure that we obtain high-quality labels and objectively measure the cost.


\subsubsection{Annotation in DataCLUE}
We use two kinds of annotation strategies in this subsection to verify the effectiveness of incorporating human in data-centric pipeline.

\textbf{Random Annotation.}
Firstly, in order to verify whether the direct use of human annotators to clean a part of the data can effectively improve the performance of the model. 
We randomly selected 2000 data samples from the data set, and then sent them to the annotation company for labeling according to the scheme described in section \ref{sec:anno_process}. Once the labeling is complete, we use the cleaned data to train a new model and measure its Marco-F1 score.

\textbf{Selective Annotation.}
Since in the baseline method we obtain an estimate of the probability of labeling error. 
Here we select the 2000 data samples that are most likely to be wrong and give them to the annotator for cleaning. 
Then evaluate the performance of the obtained data set.

\begin{table}[htbp]
\centering
\footnotesize
\begin{tabular}{ll}\toprule
Method & Macro-F1  \\ \midrule
Baseline & 72.78 \\
Random Annotation & 72.65(-0.13) \\
Selective Annotation & 74.06(+1.28) \\
Selective Annotation + Label info. & \textbf{78.36}(+5.58) \\
\bottomrule
 \hline
\end{tabular}
\caption{Analysis of Human-in-the-loop pipeline.}
\label{tab:baseline:human}
\end{table}


The results of these two strategies is reported in Table \ref{tab:baseline:human}. 
As we can see random annotation seems not perform well, the Macro-F1 score is slightly worse than the baseline. 
This may be because the random annotation does not actually correct enough wrong samples, and the correction process may cause problems such as the imbalance of the data set to be further enhanced. We will also explore this phenomenon in further work.

As for the selective annotation, it can get an improvement of 1.28 percent over baseline with the cost of \yen 700 (or \$120). This indicates that the baseline method is able to efficiently find the most likely wrong labels. But simply using the human to annotate data seems not as effective as expected. To reinforce this, we also add label information in the dataset after the selective annotation. This strategy can achieve a good result of 78.36 which is much better than using label information in the raw dataset (73.82, Analysis of Label Definition Augmentation in Table \ref{tab:baseline:text}), which indicates that adding human in the pipeline is a promising approach.

These result also shows that the DataCLUE dataset is a very challenging dataset. Relying solely on manual cleaning is not only costly, but also difficult to obtain good results (it does not exceed our baselines above). Therefore, it is very necessary to explore more and more effective methods based on DataCLUE.

\begin{CJK}{UTF8}{gbsn}
\begin{table*}[!htp]\footnotesize
\begin{center}
\begin{tabular}{ c | c | c | c }
  \hline
  ID  & Text & Original Label & Corrected  Label \\
    \hline
    60 & \makecell[c]{这个是几斤 \\  (How many catties is this)} & \makecell[c]{买家咨询活动规则 \\ (Buyer consultation activity rules)} & \makecell[c]{买家咨询商品重量 \\ (Buyers inquire about \\product weight)}  \\ 
    \hline
    353 & \makecell[c]{还有88折活动吗 \\ (Is 12\% discount still available) } & \makecell[c]{买家咨询活动规则 \\ (Buyer inquire activity rules)} & \makecell[c]{买家咨询是否有活动 \\ (Buyers inquire \\ whether there are activities)}  \\ 
    \hline
    654 & \makecell[c]{是发红包还是返回支付宝 \\ (Is it paied by red envelope \\or return to Alipay) } & \makecell[c]{买家咨询返现方式 \\ (Buyer inquire cash back method)} & \makecell[c]{买家咨询退款去向 \\ (Buyer inquire about the\\  whereabouts of refunds)}  \\ 
    \hline
    962 & \makecell[c]{新疆喀什市发不发 \\ (Can it deliver to\\ Kashgar, Xinjiang?) } & \makecell[c]{买家咨询是否可以指定快递 \\ (Buyers inquire whether express  \\delivery can be specified)} & \makecell[c]{买家咨询偏远地区是否发货 \\ (Buyers inquire whether they can\\ deliver goods in remote areas)}  \\ 
    \hline
    1090 & \makecell[c]{亲，地址可能发错了 \\ (The address may be wrong) } & \makecell[c]{买家表示商家发错地址 \\ (The buyer said the merchant\\ sent the wrong address)} & 
    \makecell[c]{买家要求修改收件信息 \\ (The buyer requested to modify\\ the receiving address)}  \\ 
    \hline
\end{tabular}
\end{center}
\caption{Label Correction Example on Baseline}
\label{tab:label_correction_example}
\end{table*}
\end{CJK}

\subsection{Training-Dynamic-Informed Label Correction}

Research has shown that there is an obvious forgetting event distribution difference between the noisy label data and the correct label data~\cite{toneva2018empirical}. The forgetting is defined as the prediction error after first time correct prediction. Inspired the observation of such training dynamics, we use the following method to correct the training data.

\begin{figure}[!htp]
	\includegraphics[width=0.5\textwidth]{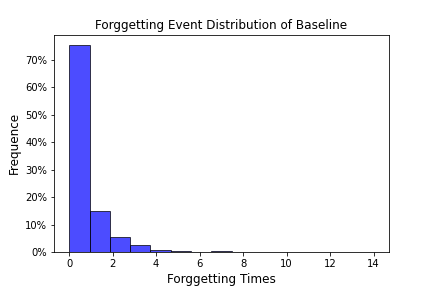}
	\caption{Forgetting Event Distribution of Baseline}
	\label{fig:baseline_jump_stat}
\end{figure}

\textbf{1. Training-Dynamic}
We record the prediction results at each epoch. We statistics the forgetting numbers of baseline model in Figure \ref{fig:baseline_jump_stat}.

Those high forgetting examples will be more likely to be mislabeled. We can use the forgetting times to filter those high-probability mislabelled data and use the most time jump targets as the corrected label.

\textbf{2. Bootstrapping Label Correction}
Bootstrapping is widely used in training of noise-robust model \cite{reed_2015_TrainingDeepNeural, wang_2021_ProSelfLCProgressiveSelf}.
With the correction or delectation of the high-probability mislabelled data, we will get a more accurate dataset and corresponding model, thus we can use the corrected dataset trained model to make predictions and correction.

\begin{figure}[!htb]
\centering
\includegraphics[width=0.46\textwidth]{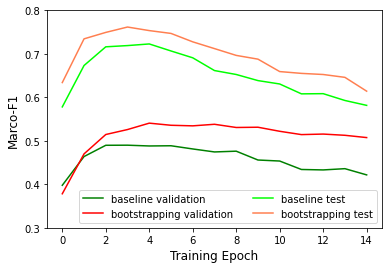}

\caption{Validation and Test Performance Gap in bootstrapping iteration 3 and baseline}\label{fig:val_test_gap}

\end{figure}

\textbf{When to Stop Bootstrapping?} 
We merged the training and validation set for label correction, and give the performance on test public. We give our label correction example in table \ref{tab:label_correction_example}, which shows both accurate correction and wrong correction.
In practical environment, we may do not have a high-quality subset to tune hyperparameters and bootstrapping iteration numbers, Figure \ref{fig:val_test_gap} illustrates the problem behind such label correction method.

\textbf{Correct or Delete?}
We can delete or correct the suspicious high forgetting times examples. We have some empirical observations: For first few epochs, correction have better performance than deletion. After several epochs, there are still some examples with high forgetting times, which can be caused by similar label definition, correction help less in this case. After 3 epochs bootstrapping correction, the performance does not increase.

From another perspective, those high forgetting training examples can also be hard examples or challenge examples, simply deleting them may lead to high information loss. We notice that correction almost gets better performance than simple deletion. How to classify those hard examples and noisy examples is an open question.



\textbf{3. Label Correction Evaluation}
We get 75.35 Marco-F1 score after bootstrapping 3 epochs. Actually, we notice a forgetting event distribution change after label correction, how to set and change the correction threshold for forgetting event times is an interesting question. The label correction method does not achieve better performance than label argumentation, which indicates that our baseline is a strong baseline and the benchmark task is challenging.

 \begin{figure}[!htp]
	\includegraphics[width=0.5\textwidth]{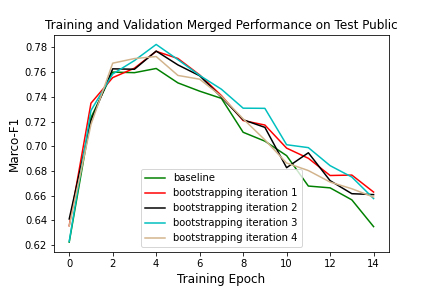}
	\caption{Merged Training Performance on Test Public, which shows more bootstrapping iterations do not provide better performance.}
	\label{fig:merged}
\end{figure}






\subsection{Discussions}
We notice that DataCLUE is a challenging dataset and human annotation does not provide much gain of performance as we originally expected. In the label correction method, there are some open questions including correcting or deleting the probability of misleading labels, how to build the validation set, and evaluating the correction method. Based on our exploration method, we discuss and give some empirical observations to those questions. These open questions give further research questions based on such benchmark and can improve the Data-Centric AI area.


\section{Conclusions}\label{conclusion} 
In this paper, we introduce DataCLUE, a benchmark and baseline for data-centric NLP. We also use three strategies to improve data quality and conduct several experimental analyses. 
Experiment results show that these strategies are simple and effective, they can achieve better performance than expensive manual label cleaning.
Meanwhile, we find the combination of different strategies do not always get improvement compared to a single strategy. It will be interesting to conduct more experiments to analyze this phenomenon. 
We will also add more tasks and baseline methods to make DataCLUE more comprehensive.
We hope that our work can promote the development of data-centric approaches in the NLP field.


\bibliography{anthology, custom}
\bibliographystyle{acl_natbib}

\clearpage
\appendix
%
\section{Other Implementations}\label{appendix}
In this section we introduce some other methods for DataCLUE.
\subsection{Model Integration Detection: Label Correction Using Cross Validation of Integrated Model}
\textbf{Motivation.}
We hypothesis the machine learning model has a certain fault tolerance rate. Then, we propose to train multiple model and use their voting to find the wrong label. 
Therefore, to save human resources, we can directly screened and replaced some of the labels in the dataset using labels predicted by the models.


\textbf{Methodology.}
We use a similar training procedure as in Section 4.1, but in each fold we run several model with different seeds.
After the training, the prediction results of data sets under different seeds will be obtained. Next, we select the data with prediction results consistent under different seeds but inconsistent with the original label. 
Then, we select a batch of data (2k+) with high confidence through the threshold. After observation, we found that these data are likely to be mislabeled data. 
Finally, we modify the original label of the data to the label predicted by the model to obtain a new data set.

\begin{table}[htbp]
\centering
\footnotesize
\begin{tabular}{ll}\toprule
Method & Macro-F1  \\ \midrule
Baseline & 72.78 \\
Cross-Validation Label Correction  & 75.40(2.62+) \\
\bottomrule
 \hline
\end{tabular}
\caption{Analysis of Label Correction by Model using Cross Validation.}
\label{tab:indep_experiment:label_correct_by_cross_validation}
\end{table}

\textbf{Results.}
We conduct experiments with the described method, the results are reported in Table~\ref{tab:indep_experiment:label_correct_by_cross_validation}. The results show that it is possible to detect mislabeled data and correct those labels. Using this simple but effective method it improves from baseline in a large margin (2.62 percent).

\end{document}